%% file: main.tex
\documentclass[10pt,twocolumn,letterpaper]{article}
\usepackage[pagenumbers]{cvpr}
\input{preamble}

\definecolor{cvprblue}{rgb}{0.21,0.49,0.74}
\usepackage[pagebackref,breaklinks,colorlinks,allcolors=cvprblue]{hyperref}
\def\confName{CVPR}
\def\confYear{2027}
\title{SCR: Detecting Backdoor-Poisoned Samples through Clean-Reference Residuals}

\author{
	Yi Yang\textsuperscript{1},
	Jinyang Huang\textsuperscript{1}\thanks{Corresponding authors: Jinyang Huang (hjy@hfut.edu.cn) and Meng Li (mengli@hfut.edu.cn).},
	Feng-Qi Cui\textsuperscript{2},
	Yu-Tong Guo\textsuperscript{1}, \\
	Sirui Zhao\textsuperscript{2},
	Haiming Jin\textsuperscript{3},
	Weihua Li\textsuperscript{4},
	Quan Bai\textsuperscript{5},
	Meng Li\textsuperscript{1}\footnotemark[1]\\
	\textsuperscript{1}Hefei University of Technology\\
	\textsuperscript{2}University of Science and Technology of China\\
	\textsuperscript{3}Shanghai Jiao Tong University\\
	\textsuperscript{4}Auckland University of Technology\\
	\textsuperscript{5}University of Tasmania\\	
}

\begin{document}
\maketitle
\input{sec/0_abstract}

\input{sec/1_intro}

\input{sec/2_related_work}

\input{sec/3_method}

\input{sec/4_experiments}

\input{sec/5_conclusion}
\typeout{SCR-BODY-PAGES=\number\numexpr\value{page}-1\relax}
\section*{LLM Usage Statement}
We used OpenAI's ChatGPT to assist with editing and refining author-produced manuscript prose, code and implementation, experimental design and analysis, and figures and diagrams. AI tools were not used for preparing related work and citations, creating or annotating datasets, or developing research ideas, hypotheses, claims, or conclusions. The authors take full responsibility for all submitted content.

{\small
\bibliographystyle{ieeenat_fullname}
\bibliography{main}
}

\end{document}

%% file: preamble.tex
\usepackage{tabularx}
\usepackage{colortbl}

%% file: sec/0_abstract.tex
\begin{abstract}
Deep neural networks are vulnerable to backdoor attacks, in which adversaries manipulate model predictions by injecting poisoned samples into training data. Detecting these samples remains challenging because legitimate semantic variation can obscure poisoning-related signals in learned representations. Consequently, class-level statistics used by existing detectors may flag benign feature deviations while overlooking poisoned samples that blend into the class distribution. In this paper, we propose Semantic Counterfactual Residualization (SCR), which uses a small set of clean reference samples to account for benign variation through sample-specific feature reconstruction. SCR identifies poisoned samples by detecting coherent deviations from these clean baselines, without requiring prior knowledge of the attacks. We evaluate SCR through experiments covering twelve dirty-label and clean-label attacks across CIFAR-10, CIFAR-100, and Tiny ImageNet. With a small clean-reference set, SCR achieves attack-averaged true positive rates above \(99.5\%\) and false positive rates below \(2\%\) on each dataset, outperforming eight representative detection baselines in macro-TPR.
\end{abstract}

%% file: sec/1_intro.tex
\section{Introduction}
\label{sec:intro}

\begin{figure}[t]
	\centering
	\includegraphics[width=\linewidth]{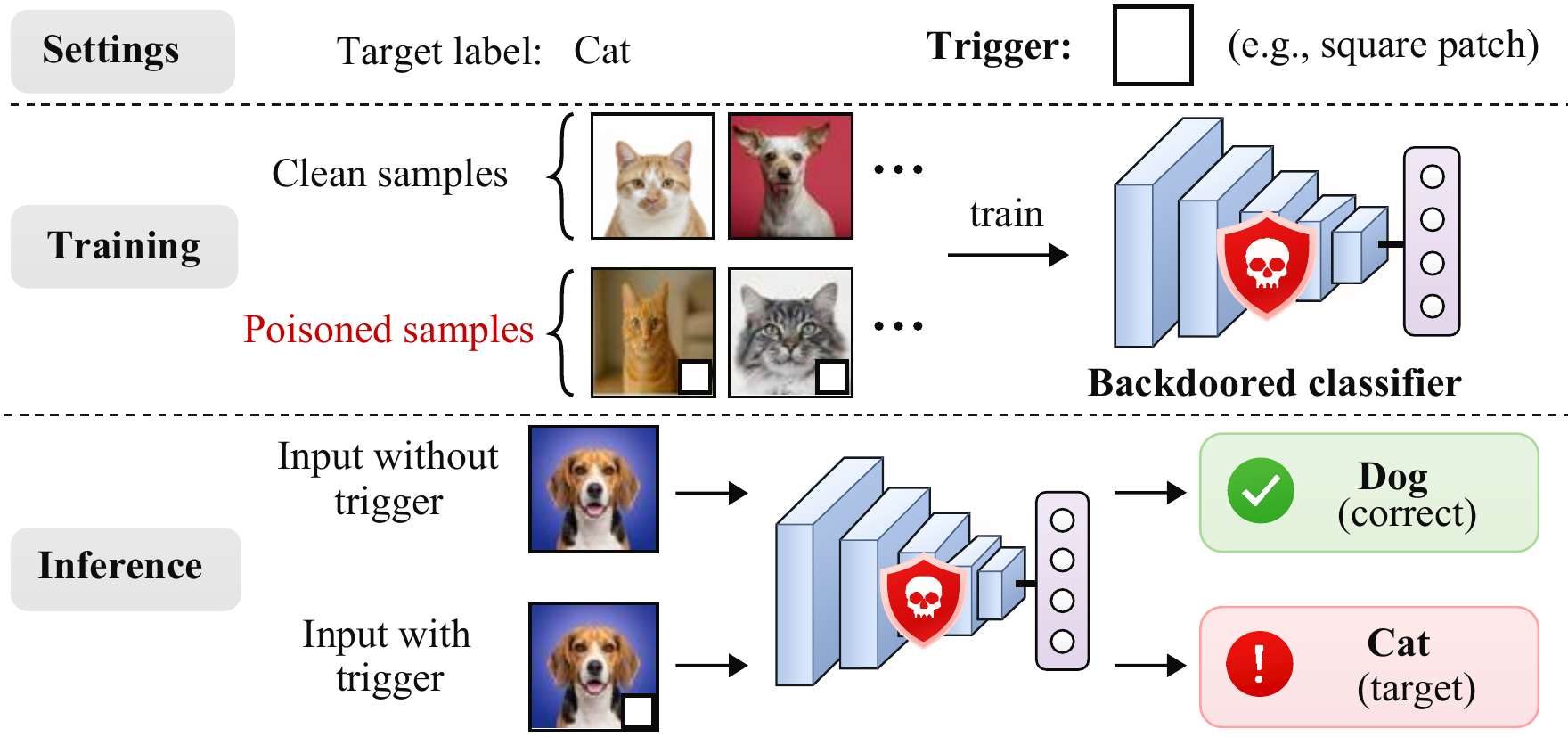}
	\caption{Illustration of a data-poisoning-based backdoor attack. Poisoned training samples containing an attacker-defined trigger induce the backdoored classifier to predict the target class on triggered inputs while maintaining correct predictions on clean inputs.}
	\label{fig:backdoor_attack}
\end{figure}

The widespread adoption of deep neural networks (DNNs) across diverse applications has raised growing concerns about their security. Among various threats, backdoor attacks pose a particularly insidious risk by injecting poisoned samples into training data to manipulate model predictions. As illustrated in Fig.~\ref{fig:backdoor_attack}, such attacks can cause backdoored models to produce attacker-desired outputs when specific triggers are present, while maintaining high accuracy on benign inputs, making them difficult to identify through conventional evaluation. To mitigate this threat, poisoned-sample detection (PSD) aims to identify contaminated training examples for subsequent dataset purification.

Existing PSD methods exploit diverse contamination signals from learned representations, activation gradients, prediction behavior, and training dynamics~\cite{tran2018spectral,tang2021demon,hayase2021spectre,ma2023beatrix,yuan2025agpd,li2025psbd,gao2025telltale}. For example,  Spectral Signatures and SPECTRE identify poisoning-related spectral abnormalities in learned representations, with SPECTRE employing robust covariance estimation and whitening~\cite{tran2018spectral,hayase2021spectre}. Other approaches leverage class-level or higher-order feature statistics~\cite{tang2021demon,ma2023beatrix}, adaptively selected multi-layer representations~\cite{hou2025flare}, or activation-gradient geometry~\cite{yuan2025agpd}.

A persistent challenge is that benign samples within the same class can exhibit substantial feature variation, potentially obscuring poisoning-related deviations. Existing methods address this issue through different statistical models. For instance, SCAn~\cite{tang2021demon} decomposes learned representations into class identity and variation components, assuming a variation distribution shared across classes. Beatrix~\cite{ma2023beatrix} instead identifies abnormal higher-order activation statistics using clean class-conditional references. Although these methods characterize benign variation, they do not explicitly remove the feature content that can be reconstructed from clean reference samples for each individual inspected sample. This motivates a complementary perspective: rather than relying solely on statistical characterization of learned representations, can we first account for the feature content of each sample that can be explained by a small set of clean references, and then examine the remaining deviations for poisoning-related structure?

To explore this idea, we consider a clean-reference baseline that adapts to each inspected representation. Within the same class, benign samples may be better explained by different combinations of clean reference features. A sample-conditioned reconstruction can therefore account for legitimate variation that a fixed class statistic may leave unexplained. If poisoning introduces feature content that cannot be adequately explained by the clean references, the corresponding deviation may persist in the reconstruction residual. However, a large residual alone does not necessarily indicate poisoning, since limited reference coverage can also leave benign variation unexplained. Rather than treating individual reconstruction errors as suspicious, we therefore seek coherent residual deviations that may arise from a shared backdoor mechanism. This motivates combining sample-conditioned clean-reference reconstruction with the detection of shared structure in the resulting residuals.

To realize this approach, we propose \emph{Semantic Counterfactual Residualization} (SCR) for PSD. Given a classifier trained on a potentially poisoned dataset and a small clean reference set, SCR optimizes a convex combination of same-class clean reference features separately for each inspected sample. The resulting counterfactual residuals are then modeled using a class-wise anchored Gaussian mixture to estimate potential shared shifts. SCR screens classes exhibiting unusually strong shifts and identifies suspicious samples through directional scoring with clean-reference-calibrated thresholds.

We evaluate SCR against eight baseline detectors across twelve backdoor attacks and three datasets. SCR achieves an macro-TPR above $99.5\%$ and an macro-FPR below $2\%$ on each dataset, outperforming the strongest baseline in macro-TPR. Ablation studies demonstrate the benefits of sample-conditioned reconstruction and coherent residual-shift detection. Furthermore, retraining after SCR-based dataset purification reduces the average attack success rate to below $1\%$ across all three datasets while maintaining overall clean accuracy within $0.3$ percentage points of Oracle purification.

Our main contributions are summarized as follows:
\begin{itemize}
	\item We introduce a \emph{sample-conditioned counterfactual residual formulation} that exposes potential poisoning-related deviations by reconstructing individualized clean-reference baselines.
	
	
	\item We develop a \emph{counterfactual residual-based detection strategy} that exploits coherent poisoning-related shifts through class-level screening and directional scoring, enabling attack-agnostic poisoned-sample detection.
	
	
	\item We demonstrate SCR's effectiveness across diverse backdoor attacks and datasets, achieving macro-TPR above $99.5\%$ with macro-FPR below $2\%$. Extensive ablations and sensitivity analyses validate its key components.
\end{itemize}

%% file: sec/2_related_work.tex
\section{Related Work}
\label{sec:related_work}

\paragraph{Backdoor attacks.}
Early backdoor attacks, such as BadNets and Blended, inject fixed patch triggers or blend predefined patterns into benign images~\cite{gu2017badnets,chen2017targeted}. Subsequent attacks introduce more diverse and stealthy trigger mechanisms. WaNet uses subtle spatial warping~\cite{nguyen2021wanet}, while SSBA generates sample-specific invisible triggers through image steganography~\cite{li2021ssba}. BPP exploits image quantization and dithering to construct imperceptible triggers~\cite{wang2022bpp}, and FTrojan~\cite{wang2022ftrojan} introduces frequency-domain perturbations by modifying discrete cosine transform coefficients. TaCT employs source-specific poisoning with auxiliary cover samples~\cite{tang2021demon}. Clean-label attacks preserve the original labels of poisoned samples. SIG embeds sinusoidal patterns into target-class images~\cite{barni2019sig}, while LC uses adversarial perturbations to suppress benign features~\cite{turner2019label}. BAAT exploits semantic attributes as sample-specific triggers~\cite{li2022baat}, whereas COMBAT alternately optimizes a trigger generator and a surrogate model to strengthen clean-label poisoning~\cite{huynh2024combat}. Adap-Blend further weakens latent separability through asymmetric trigger injection and cover samples~\cite{qi2023latent}. These attacks differ in trigger construction, label manipulation, and the resulting feature distributions, posing diverse challenges for poisoned-sample detection.

\paragraph{Backdoor defenses.}
Existing backdoor defenses include trigger reconstruction \cite{wang2019neural, tao2022better, wang2023unicorn,sun2023single}, compromised model detection and repair \cite{li2021nad, wu2021anp, zhu2023neuralpolarizer, Yang_2026_CVPR}, model inspection \cite{wang2024mm, 10.5555/3766078.3766408}, and poisoned-sample detection (PSD). Among PSD methods, Spectral Signatures identifies anomalous spectral patterns in learned representations~\cite{tran2018spectral}, while SPECTRE improves detection through robust covariance estimation and whitening~\cite{hayase2021spectre}. SCAn decomposes representations into class identity and shared variation to identify contaminated populations~\cite{tang2021demon}, whereas Beatrix detects abnormal higher-order activation statistics using Gram matrices~\cite{ma2023beatrix}. FLARE identifies discriminative feature subspaces from multi-layer activations~\cite{hou2025flare}, and AGPD exploits activation-gradient geometry for class screening and sample detection~\cite{yuan2025agpd}. Other approaches obtain detection signals through additional training or prediction behavior. Confusion Training (CT) introduces dynamically mislabeled clean samples during training to make benign samples difficult to fit while retaining the model's ability to fit backdoor poisons, allowing detection based on their different fitting behaviors~\cite{qi2023ct}. SAM-enhanced detection uses sharpness-aware minimization to amplify backdoor-related activations~\cite{zhang2026sam}, while PSBD identifies suspicious samples through dropout-induced prediction shifts~\cite{li2025psbd}. Unlike methods based on distributional statistics, activation patterns, or modified training behavior, SCR constructs a sample-conditioned clean-reference reconstruction for each inspected representation and detects coherent deviations in the resulting residual space. By combining individualized residualization with class-level screening and directional scoring, SCR provides a complementary approach to identifying poisoning-related structure.

%% file: sec/3_method.tex
\begin{figure*}[t]
	\centering
	\includegraphics[width=\linewidth]{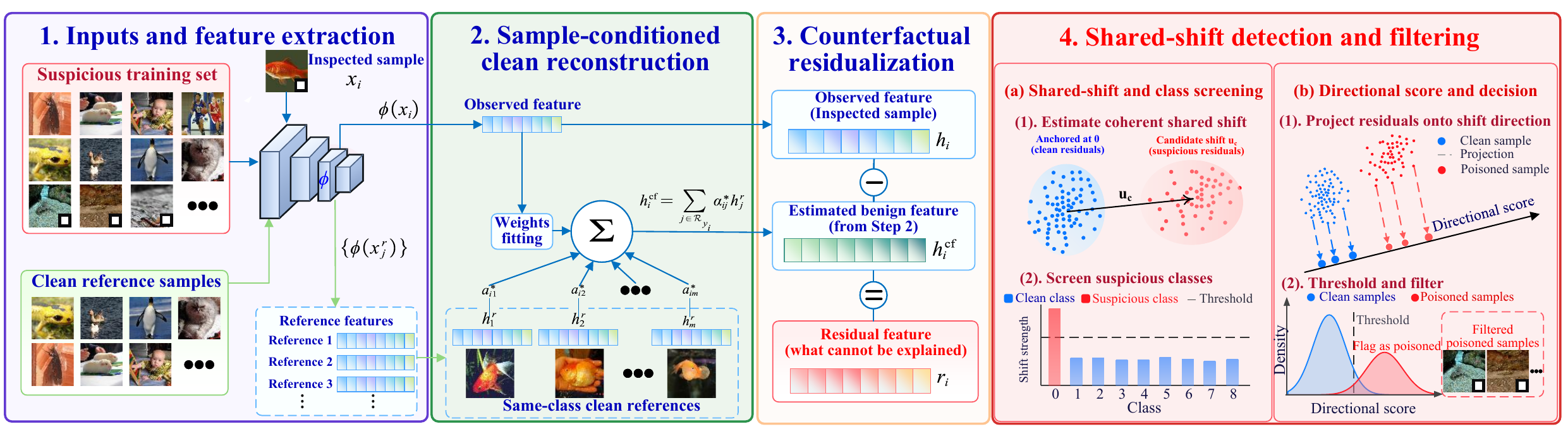}
	\caption{\textbf{Overview of SCR.}
		SCR contains four main stages:
		(1) extracting features of inspected samples and
		clean reference samples using an encoder;
		(2) reconstructing sample-specific benign representations
		from same-class clean references;
		(3) obtaining counterfactual residuals by subtracting the
		reconstructed benign components; and
		(4) estimating class-wise shared shifts, screening
		suspicious classes, and filtering poisoned samples
		through directional scoring.}
	\label{fig:overview}
\end{figure*}

\section{Threat Model}
\label{sec:threat_model}

In this section, we introduce the threat model, including the goals and capabilities of both the attacker and defender.

\textbf{Attacker.} We focus on single-target backdoor poisoning by a malicious data contributor. The attacker can modify a subset of the contributed training images and, when required by the attack, their labels. The goal is to make the trained model predict a designated target label $y_t$ on inputs that satisfy the attack's activation conditions while maintaining accuracy on benign inputs. The attacker has no control over the victim model or training procedure.

\paragraph{\textbf{Defender.}} The defender aims to detect backdoor-poisoned samples in a potentially compromised training dataset $\mathcal D_p=\{(x_i,y_i)\}_{i=1}^{N}$, where $y_i\in\{1,\ldots,C\}$ is the observed label. We assume the defender has full access to the training data and control over the model training process, including the choice of model architecture and training strategy. However, the defender has no prior knowledge of whether poisoning is present, the poisoning ratio, the attack type, trigger pattern, or the designated target class. Following common settings in prior backdoor defense work, we additionally assume access to a small set of labeled clean reference samples $\mathcal R=\{(x_j^r,y_j^r)\}_{j=1}^{M}$.

\section{Method}
\label{sec:method}


In this section, we present \emph{Semantic Counterfactual
	Residualization} (SCR), a poisoned-sample detection method
that reconstructs a sample-specific benign representation
using trusted clean references. By subtracting this estimate
from the observed representation, SCR aims to suppress
clean-explainable variation while retaining poisoning-related
deviations. It then identifies coherent residual shifts
within each observed class and uses their directions to
detect suspicious samples. The overall workflow is illustrated in Fig.~\ref{fig:overview}.

\subsection{Inputs and Feature Extraction}
\label{sec:representation}

We first train a classifier $f_\theta$ on
$\mathcal D_p$, which may have learned a backdoor from
the poisoned training data. We use $f_\theta$ to extract representations
of both inspected samples and trusted references.
Specifically, let $\phi$ denote the feature extractor
corresponding to a selected intermediate layer of
$f_\theta$. We flatten its activation to obtain
$d$-dimensional feature vectors for each $x_i$ and $x_j^r$:
\begin{equation}
	h_i=\phi(x_i) \in\mathbb R^d,\qquad
	h_j^r=\phi(x_j^r) \in\mathbb R^d.
	\label{eq:shared_features}
\end{equation}

\subsection{Sample-Conditioned Clean Reconstruction}
\label{sec:counterfactual}

Given an inspected sample $(x_i,y_i)$ from $\mathcal D_p$, SCR estimates
its benign representation using clean reference samples
from the same class $y_i$. Let
$\mathcal R_c$ denote clean reference samples belonging to class $c$, and let
$M_c=|\mathcal R_c|$, SCR reconstructs each inspected
representation through a sample-specific combination
of same-class reference features. Specifically,
we determine the reconstruction coefficients $\boldsymbol{\alpha}_i^*$ by solving
\begin{equation}
	\begin{aligned}
		\boldsymbol{\alpha}_i^*
		=\arg\min_{\boldsymbol{\alpha}}\quad &
		\left\|
		h_i-\sum_{j\in\mathcal R_{y_i}}
		\alpha_{ij}h_j^r
		\right\|_2^2
		+\lambda\|\boldsymbol{\alpha}\|_2^2,\\
		\text{s.t.}\quad &
		\alpha_{ij}\geq 0,\qquad
		\sum_{j\in\mathcal R_{y_i}}\alpha_{ij}=1,
	\end{aligned}
	\label{eq:simplex_reconstruction}
\end{equation}
where $\lambda\geq0$ controls ridge regularization.
We solve this convex optimization problem using
projected gradient descent with Euclidean simplex
projection.

The simplex constraints ensure that the reconstructed
representation is a convex combination of clean reference
features. Importantly, the coefficients are optimized
independently for each inspected sample, allowing the
reconstruction to adapt to its individual feature
characteristics. Given the optimized $\boldsymbol{\alpha}_i^*$, the reconstructed counterfactual representation $h_i^{cf}$ of the inspected sample $x_i$ is
\begin{equation}
	h_i^{\mathrm{cf}}
	=\sum_{j\in\mathcal R_{y_i}}
	\alpha_{ij}^*h_j^r.
	\label{eq:counterfactual}
\end{equation}
Thus, $h_i^{\mathrm{cf}}$ provides a sample-specific
estimate of the benign component explainable by
the clean reference samples.

\subsection{Counterfactual Residualization}
\label{sec:residualization}

Having estimated the benign counterpart of each
inspected sample, SCR isolates the variation not
explained by the reconstruction. Specifically,
we define the counterfactual residual as
\begin{equation}
	r_i=h_i-h_i^{\mathrm{cf}},
	\label{eq:counterfactual_residual}
\end{equation}
where $r_i\in\mathbb R^d$ has the same dimensionality
as the observed and reconstructed representations. The residual captures the discrepancy between the
observed feature and its reference-conditioned
benign estimate. When the reconstruction accounts
for benign semantic variation without reproducing
the poisoning-related component, the latter can
become more prominent in the residual space.

Nevertheless, large residuals are not necessarily
indicative of poisoning, since legitimate
sample-specific variation may remain unexplained.
SCR therefore examines whether residuals exhibit
a coherent shift shared by multiple samples,
rather than identifying poisoned samples directly
from their residual magnitudes.

\subsection{Shared-Shift Detection and Filtering}
\label{sec:shared_shift_detection}

Given the counterfactual residuals, SCR performs
detection in two steps. It first estimates coherent
residual shifts within each observed class and
screens classes exhibiting anomalous shift strengths.
For the screened classes, it then identifies
individual suspicious samples through directional
scoring and trusted-reference thresholding.

\paragraph{Class-wise shared-shift estimation.}

Samples affected by the same poisoning mechanism
may exhibit a common deviation in residual space.
To capture this structure, SCR models the residual
distribution separately for each observed class. Let $\mathcal I_c=\{i:y_i=c\}$ denote the indices
of inspected samples of class $c$, and
let $n_c=|\mathcal I_c|$.
We adopt an anchored two-component Gaussian
mixture model:
\begin{equation}
	p(r\mid c)
	=(1-\pi_c)\mathcal N(r;0,I_d)
	+\pi_c\mathcal N(r;u_c,I_d),
	\label{eq:mixture}
\end{equation}
where $\pi_c\in(0,1)$ is the mixing proportion
of the candidate shifted component, and
$u_c\in\mathbb R^d$ denotes its residual
shift.

The component anchored at zero represents the
residual distribution expected after removing
clean-explainable variation, while the shifted
component captures coherent deviations from this
reference point. The identity covariance assumption
provides a simple model for estimating
the shift.

We estimate \(\pi_c\) and \(\mathbf{u}_c\) from the residuals of class \(c\) using expectation-maximization (EM), with the mixture proportion constrained to \([\pi_{\min},\pi_{\max}]\). Details of EM estimation are provided in Appendix~\ref{app:scr_configuration}.

\paragraph{Class-level screening.}

Shared shifts can arise from benign variation.
Under the single-target setting defined in our threat model (Sec.~\ref{sec:threat_model}),
a poisoning-related shift may stand out relative to those of other classes.
SCR measures the normalized shift strength of each class $c$
as $U_c=\frac{\|u_c\|_2}{\sqrt d}$ and computes a standardized
score using the median and median absolute deviation (MAD) across all classes.
\begin{equation}
	Z_c=
	\frac{U_c-\widetilde U}
	{\max\{\operatorname{MAD}_U,
		\epsilon_{\mathrm{MAD}}\}},
	\label{eq:class_score}
\end{equation}
where
$\widetilde U=\operatorname{median}_c U_c$,
$\operatorname{MAD}_U
=\operatorname{median}_c|U_c-\widetilde U|$,
and $\epsilon_{\mathrm{MAD}}>0$ prevents numerical
instability. SCR screens each class using
$Z_c>\tau_{\mathrm{cls}}$, where
$\tau_{\mathrm{cls}}>0$ controls the sensitivity
of class-level screening. Every class that passes this criterion
proceeds to sample-level detection.

\paragraph{Directional sample scoring.}
For each screened class, SCR identifies suspicious samples
by measuring their residual shifts along the estimated
shared direction. Specifically, for a class $c$ with
$\|u_c\|_2>0$, we define
\begin{equation}
	\hat u_c=\frac{u_c}{\|u_c\|_2},
	\qquad
	s_i=\hat u_c^\top r_i,
	\quad i\in\mathcal I_c.
	\label{eq:sample_score}
\end{equation}
Unlike residual magnitude, which aggregates deviations
across all directions, directional scoring focuses on
the component associated with the estimated shared
shift. A larger score indicates stronger evidence
of belonging to the candidate shifted component.
Under the Gaussian mixture in Eq.~\eqref{eq:mixture},
the component log-likelihood ratio is monotonic
in $s_i$, providing a statistical basis for this
scoring strategy.

\paragraph{Clean-reference thresholding and filtering.}
To translate directional scores into sample-level
decisions, SCR establishes a class-specific threshold
using clean reference samples. This provides a
benign score baseline along the estimated shift
direction without requiring knowledge of the
poisoning rate. We first compute leave-one-out residuals for clean
references to avoid self-reconstruction. For each clean
reference $j$, we solve
Eq.~\eqref{eq:simplex_reconstruction} after excluding
$j$ from its reference set, yielding
\begin{equation}
	r_j^r=h_j^r-
	\sum_{k\in\mathcal R_{y_j^r}\setminus\{j\}}
	\alpha_{jk}^{r,*}h_k^r,
	\label{eq:reference_loo}
\end{equation}
where $\boldsymbol{\alpha}_j^{r,*}$ is the
corresponding reconstruction coefficients.

For each screened class $c$, we project the
reference residuals onto its estimated shift
direction:
\begin{equation}
	s_{jc}^r=\hat u_c^\top r_j^r,
	\qquad j\in\mathcal R_c.
	\label{eq:reference_score}
\end{equation}
Let $m_c$ denote the number of reference
scores, and $\bar s_c^r$ and $\hat\sigma_c^r$
be their sample mean and standard deviation. For $m_c\geq2$, we adopt a one-sided predictive
Student-$t$ threshold:
\begin{equation}
	\tau_c=\bar s_c^r+
	t_{1-\beta,m_c-1}\hat\sigma_c^r
	\sqrt{1+\frac{1}{m_c}},
	\label{eq:student_threshold}
\end{equation}
where $t_{1-\beta,m_c-1}$ is the
$(1-\beta)$-quantile of the Student-$t$
distribution with $m_c-1$ degrees of freedom,
and $\beta$ specifies the nominal upper-tail
significance level. SCR flags samples with $s_i>\tau_c$
in screened classes as potentially poisoned,
while retaining all samples in unscreened classes.

%% file: sec/4_experiments.tex
\section{Experiments}
\label{sec:experiments}

We evaluate SCR's effectiveness in detecting poisoned samples across three datasets and twelve dirty-label and clean-label attacks. We first describe the experimental setup and compare SCR with eight baseline detectors, followed by a case study illustrating its detection workflow. We then examine its key design choices through ablations and sensitivity analyses, and evaluate backdoor suppression and clean accuracy after dataset purification and retraining.

\subsection{Experimental setup}
\label{sec:setup}
\paragraph{Datasets, model and hyperparameters.} We conduct experiments on CIFAR-10, CIFAR-100~\cite{krizhevsky2009learning}, and Tiny ImageNet~\cite{deng2009imagenet}, using ResNet-18~\cite{he2016deep} as the default architecture. Note that in our threat model, the defender has full control on the training process and thus free to choose any architecture, and the sole goal of the defender is to detect and filter the potential backdoor data that may exist in the potential poisoned dataset. By default, we set $\tau_{cls}=10$, $\beta=0.01$ and use $layer4$ of ResNet-18 to extract features from samples. Appendix A provides dataset statistics, feature dimensions, and implementation details.

\paragraph{Attacks and poisoning rates.}
We evaluate twelve representative backdoor attacks, including eight dirty-label attacks (BadNets~\cite{gu2017badnets}, Blended~\cite{chen2017targeted}, BPP~\cite{wang2022bpp}, FTrojan~\cite{wang2022ftrojan}, SSBA~\cite{li2021ssba}, WaNet~\cite{nguyen2021wanet}, Adap-Blend~\cite{qi2023latent}, and TaCT~\cite{tang2021demon}) and four clean-label attacks (BAAT~\cite{li2022baat}, COMBAT~\cite{huynh2024combat}, LC~\cite{turner2019label}, and SIG~\cite{barni2019sig}). In the main comparison, we focus on effective backdoor configurations with ASR $>80\%$ and small to moderate poisoning rates (typically $1\%$ or $5\%$). On CIFAR-100 and Tiny ImageNet, COMBAT, LC, and SIG are excluded because their attempted configurations did not meet the ASR criterion. Table~\ref{tab:main_detection} reports the poisoning rates for each evaluated configuration, while Appendix A.3 provides detailed attack configurations and undefended clean-test accuracy (ACC) and ASR. For dirty-label attacks, we report the global poisoning rate, defined as the proportion of poisoned samples in the entire training dataset. For attacks involving cover samples (Adap-Blend, TaCT, and WaNet), we additionally report the cover rate. For clean-label attacks, we report both the global poisoning rate and the target-class poisoning rate, defined as the proportion of poisoned samples within the target class. Appendix C presents additional comparisons across six dirty-label attacks at global poisoning rates of $5\%$ and $10\%$. We also evaluate an adaptive attack designed for SCR, with results reported in Appendix E.

\paragraph{Compared detectors} We compare SCR with eight classic and state-of-the-art backdoor poisoned sample detection methods, including PSBD~\cite{li2025psbd}, SCAn~\cite{tang2021demon}, Beatrix~\cite{ma2023beatrix}, FLARE\textsuperscript{*}~\cite{hou2025flare}, AGPD~\cite{yuan2025agpd}, CT~\cite{qi2023ct}, SPECTRE\textsuperscript{*}~\cite{hayase2021spectre} and SAM-Beatrix~\cite{zhang2026sam}. Note that we evaluate SPECTRE in a semi-oracle model assuming it knows the target class and the number of poisoned samples. All method has the access to the poisoned training datasets to inspect and the trained model. Methods annoateted with * do not require clean samples to perform detection, while other methods like SCR require clean samples to perform detection, by default, we use 30 samples per class for all methods.

\paragraph{Detection metrics.}
We evaluate detection performance using the true positive rate (TPR) and false positive rate (FPR), where higher TPR and lower FPR indicate better performance. For each experimental setting, we conduct 10 independent runs for all methods and report the mean results. Since the observed variations across runs are small, we omit the variance for clarity. Dataset-level averages are computed by assigning equal weight to each evaluated attack. We additionally report SCR's precision and F1 scores under the same settings as Table~\ref{tab:main_detection}, along with an analysis of their implications and trade-offs, in Appendix D.

\begin{table*}[t]
	\caption{Poisoned-sample detection, reported as TPR/FPR (\%; $\uparrow$TPR, $\downarrow$FPR). Attack annotations give the global poisoning rate, followed by $c$ (global cover rate) or $t$ (target-class poisoning rate for clean-label attacks), when applicable.  SAM-B.\ denotes SAM-Beatrix. Bold TPR/FPR pairs indicate the highest TPR in each attack or Average row, including all ties at the displayed precision.  SCAn's -/- indicates failure to correctly identify the poisoned class. Gray text denotes failed detection ($\mathrm{TPR}<80\%$).}
	\label{tab:main_detection}
	\centering\scriptsize
	\definecolor{scrFailed}{gray}{0.40}
	\definecolor{scrAverage}{gray}{0.92}
	\setlength{\tabcolsep}{3pt}
	\begin{tabularx}{\textwidth}{@{}l*{9}{>{\centering\arraybackslash}X}@{}}
		\toprule
		Attack & PSBD & SCAn & Beatrix & FLARE & AGPD & CT & SPECTRE & SAM-B. & SCR \\
		\midrule
		\multicolumn{10}{l}{\textbf{CIFAR-10}} \\
		BadNets $(1\%)$ & \textbf{100.00/11.31} & \textcolor{scrFailed}{-/-} & \textcolor{scrFailed}{62.40/3.92} & \textcolor{scrFailed}{7.00/0.20} & \textcolor{scrFailed}{70.80/0.00} & 95.20/4.00 & 94.80/0.05 & 96.60/5.01 & 98.20/1.02 \\
		Blended $(1\%)$ & 91.80/21.30 & 96.40/0.00 & 96.20/4.74 & 97.80/0.00 & 97.80/0.00 & \textbf{100.00/0.99} & 99.80/0.00 & \textcolor{scrFailed}{72.80/5.01} & \textbf{100.00/1.12} \\
		BPP $(1\%)$ & 96.80/22.80 & \textcolor{scrFailed}{-/-} & 98.40/4.11 & \textcolor{scrFailed}{2.00/9.77} & 97.00/0.00 & \textcolor{scrFailed}{0.00/1.56} & 98.00/0.02 & 84.94/5.01 & \textbf{100.00/0.56} \\
		FTrojan $(5\%)$ & \textbf{100.00/15.66} & \textbf{100.00/0.00} & \textbf{100.00/3.66} & 99.68/0.00 & 98.80/0.00 & \textbf{100.00/0.92} & 95.28/0.25 & \textcolor{scrFailed}{64.36/5.01} & \textbf{100.00/2.12} \\
		SSBA $(5\%)$ & \textbf{100.00/13.53} & 95.88/0.00 & 87.88/3.23 & \textcolor{scrFailed}{0.88/49.66} & 98.40/0.02 & 98.24/1.93 & 80.40/1.03 & 99.48/5.01 & \textbf{100.00/1.37} \\
		Adap-Blend $(1\%;\ c{=}1\%)$ & 90.40/21.51 & \textcolor{scrFailed}{-/-} & \textcolor{scrFailed}{40.80/3.72} & \textcolor{scrFailed}{1.00/9.90} & 97.00/0.17 & \textcolor{scrFailed}{0.00/5.57} & 83.40/0.17 & \textcolor{scrFailed}{64.00/5.01} & \textbf{98.60/3.57} \\
		TaCT $(2\%;\ c{=}3\%)$ & 98.20/17.14 & \textcolor{scrFailed}{-/-} & \textcolor{scrFailed}{46.50/2.74} & 88.10/2.73 & 95.20/0.00 & 82.70/4.98 & 91.80/0.17 & 91.00/5.01 & \textbf{99.80/5.94} \\
		WaNet $(5\%;\ c{=}10\%)$ & 98.96/22.69 & 90.00/0.00 & \textcolor{scrFailed}{58.36/4.56} & \textcolor{scrFailed}{1.40/9.32} & 99.24/0.03 & 96.84/3.16 & \textcolor{scrFailed}{53.64/2.44} & \textcolor{scrFailed}{63.65/5.00} & \textbf{99.92/1.37} \\
		BAAT $(1\%;\ t{=}10\%)$ & \textcolor{scrFailed}{0.20/26.38} & \textcolor{scrFailed}{-/-} & 89.60/5.10 & \textcolor{scrFailed}{0.00/20.32} & \textcolor{scrFailed}{0.00/3.33} & \textbf{100.00/4.89} & 97.00/0.00 & 90.60/5.01 & \textbf{100.00/2.44} \\
		COMBAT $(5\%;\ t{=}50\%)$ & \textcolor{scrFailed}{14.08/25.79} & 97.56/0.00 & 94.52/4.47 & \textcolor{scrFailed}{0.16/21.14} & \textcolor{scrFailed}{0.00/18.17} & 90.80/1.68 & \textcolor{scrFailed}{37.52/32.90} & 95.56/5.01 & \textbf{99.96/0.74} \\
		LC $(1\%;\ t{=}10\%)$ & 95.80/21.89 & \textcolor{scrFailed}{-/-} & \textbf{100.00/3.57} & \textbf{100.00/0.00} & \textcolor{scrFailed}{0.00/15.69} & \textbf{100.00/1.00} & \textbf{100.00/0.00} & \textcolor{scrFailed}{75.00/5.01} & \textbf{100.00/1.52} \\
		SIG $(1\%;\ t{=}10\%)$ & \textcolor{scrFailed}{2.60/19.28} & \textcolor{scrFailed}{-/-} & 99.40/2.68 & 97.00/0.00 & \textbf{100.00/11.57} & 99.40/3.34 & 99.80/0.00 & \textbf{100.00/5.01} & \textbf{100.00/1.72} \\
		\rowcolor{scrAverage}
		\textbf{Average} & \textcolor{scrFailed}{74.07/19.94} & \textcolor{scrFailed}{39.99/0.00} & 81.17/3.88 & \textcolor{scrFailed}{41.25/10.25} & \textcolor{scrFailed}{71.19/4.08} & 80.27/2.84 & 85.95/3.09 & 83.17/5.01 & \textbf{99.71/1.96} \\
		\midrule
		\multicolumn{10}{l}{\textbf{CIFAR-100}} \\
		BadNets $(5\%)$ & \textbf{100.00/1.00} & 95.60/0.00 & \textcolor{scrFailed}{26.48/3.63} & 88.12/0.00 & \textcolor{scrFailed}{1.48/0.11} & \textcolor{scrFailed}{77.76/0.97} & 82.84/0.90 & \textcolor{scrFailed}{43.84/5.07} & 99.40/0.52 \\
		Blended $(1\%)$ & \textcolor{scrFailed}{26.40/1.19} & 99.60/0.00 & \textcolor{scrFailed}{76.80/3.60} & 92.00/1.82 & 99.40/0.00 & \textbf{100.00/1.06} & \textcolor{scrFailed}{20.20/0.81} & 85.20/5.05 & \textbf{100.00/0.62} \\
		BPP $(1\%)$ & 97.80/5.31 & 98.80/0.00 & \textcolor{scrFailed}{2.60/3.20} & 94.00/0.01 & 94.80/0.00 & \textcolor{scrFailed}{0.00/2.20} & 93.00/0.07 & 93.54/5.04 & \textbf{100.00/0.26} \\
		FTrojan $(5\%)$ & 99.88/1.41 & 99.80/0.00 & 99.20/3.94 & 99.88/0.02 & \textcolor{scrFailed}{0.00/0.03} & \textcolor{scrFailed}{0.00/0.06} & 81.68/0.96 & 99.20/5.07 & \textbf{100.00/0.22} \\
		SSBA $(5\%)$ & \textbf{100.00/1.00} & 99.60/0.00 & \textcolor{scrFailed}{6.32/3.33} & \textbf{100.00/4.81} & \textcolor{scrFailed}{0.00/0.05} & 88.00/0.00 & 82.88/0.90 & \textcolor{scrFailed}{71.16/5.07} & \textbf{100.00/0.44} \\
		Adap-Blend $(1\%;\ c{=}1\%)$ & \textcolor{scrFailed}{9.60/1.21} & \textcolor{scrFailed}{-/-} & \textcolor{scrFailed}{13.60/3.81} & \textcolor{scrFailed}{0.20/0.98} & 94.20/1.08 & 98.60/0.41 & \textcolor{scrFailed}{17.20/0.84} & \textcolor{scrFailed}{78.60/5.05} & \textbf{99.80/0.62} \\
		TaCT $(0.5\%;\ c{=}0.75\%)$ & \textcolor{scrFailed}{2.80/1.39} & \textcolor{scrFailed}{-/-} & \textcolor{scrFailed}{3.20/4.21} & \textcolor{scrFailed}{0.00/0.10} & \textcolor{scrFailed}{0.00/0.06} & 95.20/1.50 & 85.20/0.07 & \textcolor{scrFailed}{17.60/5.01} & \textbf{100.00/0.98} \\
		WaNet $(1\%;\ c{=}2\%)$ & \textcolor{scrFailed}{56.40/8.09} & 99.80/0.00 & \textcolor{scrFailed}{14.40/4.10} & \textcolor{scrFailed}{0.00/1.02} & \textbf{100.00/0.13} & \textcolor{scrFailed}{0.00/3.82} & \textcolor{scrFailed}{27.20/0.74} & \textcolor{scrFailed}{32.23/5.04} & 99.60/0.13 \\
		BAAT $(0.1\%;\ t{=}10\%)$ & \textcolor{scrFailed}{0.00/1.58} & \textcolor{scrFailed}{-/-} & 94.00/3.79 & \textcolor{scrFailed}{0.00/1.00} & \textcolor{scrFailed}{0.00/0.15} & 98.00/1.80 & \textbf{100.00/0.00} & 98.00/5.01 & \textbf{100.00/0.11} \\
		\rowcolor{scrAverage}
		\textbf{Average} & \textcolor{scrFailed}{54.76/2.46} & \textcolor{scrFailed}{65.91/0.00} & \textcolor{scrFailed}{37.40/3.73} & \textcolor{scrFailed}{52.69/1.08} & \textcolor{scrFailed}{43.32/0.18} & \textcolor{scrFailed}{61.95/1.31} & \textcolor{scrFailed}{65.58/0.59} & \textcolor{scrFailed}{68.82/5.05} & \textbf{99.87/0.43} \\
		\midrule
		\multicolumn{10}{l}{\textbf{Tiny ImageNet}} \\
		BadNets $(1\%)$ & 93.80/11.45 & 99.80/0.00 & 99.90/51.50 & \textbf{100.00/0.00} & 99.70/0.27 & \textbf{100.00/0.67} & \textcolor{scrFailed}{50.80/0.50} & 99.80/5.05 & \textbf{100.00/0.06} \\
		Blended $(1\%)$ & 83.40/3.56 & 96.70/0.00 & \textcolor{scrFailed}{48.10/5.50} & \textcolor{scrFailed}{0.10/0.50} & \textcolor{scrFailed}{0.00/0.18} & 99.80/0.62 & \textcolor{scrFailed}{50.80/0.50} & \textcolor{scrFailed}{68.70/5.05} & \textbf{100.00/0.17} \\
		BPP $(1\%)$ & 96.60/3.56 & 99.80/0.00 & 99.90/5.35 & 99.40/0.00 & \textcolor{scrFailed}{0.00/0.01} & 99.10/0.75 & 82.00/0.18 & 97.81/5.04 & \textbf{100.00/0.12} \\
		FTrojan $(5\%)$ & 99.98/3.33 & 99.96/0.00 & \textbf{100.00/5.10} & 99.74/0.02 & 99.80/35.00 & \textbf{100.00/0.47} & 90.70/0.49 & \textbf{100.00/5.07} & \textbf{100.00/0.12} \\
		SSBA $(1\%)$ & 86.70/3.05 & 97.10/0.00 & 87.90/5.62 & \textcolor{scrFailed}{0.00/0.10} & \textcolor{scrFailed}{0.00/0.08} & \textbf{100.00/1.20} & \textcolor{scrFailed}{67.60/0.33} & 84.70/5.05 & 99.90/0.50 \\
		Adap-Blend $(0.3\%;\ c{=}0.3\%)$ & \textcolor{scrFailed}{3.00/3.60} & \textcolor{scrFailed}{-/-} & \textcolor{scrFailed}{7.00/5.32} & \textcolor{scrFailed}{0.00/0.49} & 97.67/33.43 & 85.00/3.92 & \textcolor{scrFailed}{56.00/0.13} & \textcolor{scrFailed}{29.33/5.02} & \textbf{98.67/0.30} \\
		TaCT $(0.2\%;\ c{=}0.3\%)$ & \textcolor{scrFailed}{9.00/3.41} & \textcolor{scrFailed}{-/-} & \textcolor{scrFailed}{35.00/5.21} & \textcolor{scrFailed}{0.00/0.50} & 98.00/36.00 & 99.50/1.58 & 89.00/0.02 & \textcolor{scrFailed}{79.00/5.00} & \textbf{100.00/0.56} \\
		WaNet $(1\%;\ c{=}2\%)$ & \textcolor{scrFailed}{4.30/1.97} & 98.20/0.00 & \textcolor{scrFailed}{5.80/6.14} & \textcolor{scrFailed}{0.10/0.15} & 83.40/0.03 & 99.60/0.22 & \textcolor{scrFailed}{67.90/0.32} & \textcolor{scrFailed}{51.02/5.04} & \textbf{99.70/0.01} \\
		BAAT $(0.25\%;\ t{=}50\%)$ & \textcolor{scrFailed}{0.40/2.77} & \textcolor{scrFailed}{-/-} & \textcolor{scrFailed}{0.00/5.25} & \textcolor{scrFailed}{0.00/0.50} & 96.80/36.93 & \textcolor{scrFailed}{0.00/1.66} & \textcolor{scrFailed}{40.40/0.15} & 90.00/5.00 & \textbf{100.00/0.27} \\
		\rowcolor{scrAverage}
		\textbf{Average} & \textcolor{scrFailed}{53.02/4.08} & \textcolor{scrFailed}{65.73/0.00} & \textcolor{scrFailed}{53.73/10.55} & \textcolor{scrFailed}{33.26/0.25} & \textcolor{scrFailed}{63.93/15.77} & 87.00/1.23 & \textcolor{scrFailed}{66.13/0.29} & \textcolor{scrFailed}{77.82/5.04} & \textbf{99.81/0.23} \\
		\bottomrule
	\end{tabularx}
\end{table*}

\subsection{Effectiveness of SCR}
\label{sec:comparison}
As shown in Table~\ref{tab:main_detection}, SCR detects at least $98\%$ of poisoned samples in every evaluated setting and achieves the highest mean TPR across all three datasets, with a mean FPR below $2\%$ in each case. In contrast, every baseline fails to effectively detect poisoned samples under at least one attack on each dataset. The largest improvement over the strongest baseline is observed on CIFAR-100, where SCR outperforms SAM-Beatrix by $31$ percentage points in  macro-TPR while achieving a lower macro-FPR. On Tiny ImageNet, CT achieves the highest macro-TPR among the baselines but fails to detect any BAAT poisoned samples, whereas SCR detects all of them. Under Adap-Blend, SCR also achieves a higher TPR than AGPD while maintaining a substantially lower FPR ($0.30\%$ versus $33.43\%$). These results demonstrate that SCR maintains high detection rates and low false-positive rates across diverse attacks, supporting its effectiveness in screening potentially poisoned training datasets.

We further evaluate SCR on the original, unpoisoned training datasets. In all three datasets, SCR identifies no suspicious classes and consequently removes no clean samples. This result demonstrates that SCR can preserve clean training data in the absence of poisoning, further supporting the effectiveness of its class screening mechanism.

\begin{figure*}[!t]
	\centering
	\subfloat[Raw feature]{%
		\includegraphics[height=0.24\textwidth]{
			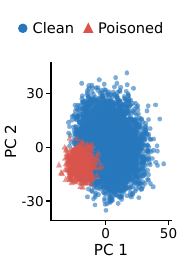
	}}\hspace{0.01\textwidth}%
	\subfloat[Reconstruction]{%
		\includegraphics[height=0.24\textwidth]{
			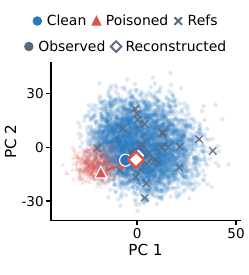
	}}\hspace{0.01\textwidth}%
	\subfloat[Residual feature]{%
		\includegraphics[height=0.24\textwidth]{
			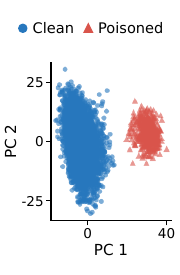
	}}\hspace{0.01\textwidth}%
	\subfloat[Class-level screening]{%
		\includegraphics[height=0.24\textwidth]{
			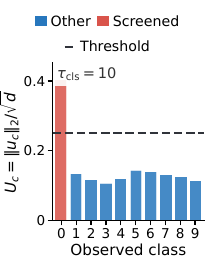
	}}\hspace{0.01\textwidth}%
	\subfloat[Directional score dist]{%
		\includegraphics[height=0.24\textwidth]{
			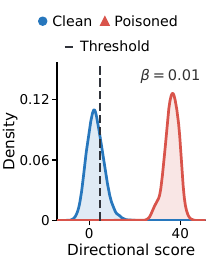
	}}
	\caption{A case study of the SCR workflow on CIFAR-10 under BAAT with a 1\% poisoning rate.}
	\label{fig:case_study}
\end{figure*}

\subsection{Case Study}
\label{sec:case_study}

We use BAAT on CIFAR-10 with a global poisoning rate
of $1\%$ ($10\%$ within the target class) as a case
study to illustrate how SCR reveals poisoning-related
patterns and identifies suspicious samples.
Fig.~\ref{fig:case_study} presents the intermediate
results. Principal Component Analysis (PCA) is used
solely to project high-dimensional features into
two dimensions for visualization, without affecting
the detection process.

Fig.~\ref{fig:case_study} (a) shows that clean and
poisoned samples overlap in the raw
feature space, making them difficult to distinguish.
In Fig.~\ref{fig:case_study} (b), we visualize the sample-specific
reconstructions of a clean and a poisoned sample,
respectively. By reconstructing each sample from
same-class clean references, SCR accounts for
sample-specific benign variation and isolates
unexplained deviations. Consequently, the residual
feature space in Fig.~\ref{fig:case_study} (c) exhibits clearer
separation between clean and poisoned samples,
illustrating the effectiveness of counterfactual
residualization.

Fig.~\ref{fig:case_study} (d) illustrates class-level screening based
on the estimated shared-shift strength. Only the
attacked class exceeds the screening threshold,
allowing SCR to focus on suspicious classes while
avoiding unnecessary filtering of other classes.
Finally, Fig.~\ref{fig:case_study} (e) presents the directional score
distributions within the screened class. Poisoned
samples generally exhibit higher scores than clean
samples, enabling their identification using a
threshold calibrated from trusted clean references.
Together, these results demonstrate how SCR combines
sample-specific residualization, class-level
screening, and directional scoring to distinguish
poisoned samples from benign variation.

\subsection{Ablation Studies}
\label{sec:ablation}
We ablate SCR's residualization strategy, class-level screening, and
directional sample scoring (Table~\ref{tab:ablation_preview}).
Appendix B.1 compares different Gaussian mixture
parameterizations.

\begin{table}[t]
\centering\scriptsize
\caption{Macro-TPR/FPR (\%) results of ablation studies.}
\label{tab:ablation_preview}
\setlength{\tabcolsep}{2pt}
\renewcommand{\arraystretch}{1.08}
\begin{tabular*}{\columnwidth}{@{\extracolsep{\fill}}lccc@{}}
\toprule
Configuration & CIFAR-10 & CIFAR-100 & Tiny ImageNet \\
\midrule
\textbf{Full SCR} & 99.71/1.96 & 99.87/0.43 & 99.81/0.23 \\
\midrule
\multicolumn{4}{l}{\textit{Residualization}} \\
\quad Class Mean & 90.51/0.58 & 77.48/0.01 & 44.44/0.04 \\
\quad Nearest Neighbor & 18.13/0.16 & 0.00/0.00 & 0.00/0.00 \\
\quad kNN Mean & 80.08/0.37 & 88.36/0.01 & 65.56/0.04 \\
\midrule
\multicolumn{4}{l}{\textit{Class screening}} \\
\quad Without Gate & 99.81/47.07 & 99.87/56.22 & 99.81/63.03 \\
\midrule
\multicolumn{4}{l}{\textit{Sample scoring}} \\
\quad Magnitude & 70.23/0.72 & 15.97/0.02 & 25.48/0.03 \\
\bottomrule
\end{tabular*}
\end{table}

\paragraph{Residualization strategies.}
We compare SCR's optimized convex reconstruction with three alternative reconstruction choices: \emph{Class Mean} uses the
mean of all clean references, \emph{Nearest Neighbor} uses the closest
reference, and \emph{kNN Mean} uniformly averages the five closest
references. All variants retain the same downstream design components.
Nearest Neighbor yields only $18.13\%$ TPR on CIFAR-10 and no
detections on CIFAR-100 or Tiny ImageNet. Averaging references
substantially improves TPR, although neither averaging rule
consistently dominates the other: Class Mean performs better on
CIFAR-10, while kNN Mean performs better on the other two datasets.
SCR achieves the highest TPR on all datasets, accompanied by
slightly higher FPR. These results support
adapting the reconstruction weights to each sample rather than
relying on a single reference or fixed averaging weights.

\paragraph{Class-level screening.}
By default, SCR first performs class-level screening to identify suspicious classes before applying sample-level filtering. Removing the class gate in Eq.~(\ref{eq:class_score}) applies sample-level filtering to every class, including clean ones. Although trusted references and inspected samples in a clean class follow the same underlying distribution, the estimated shift direction is fitted adaptively from the inspected residuals and may capture incidental structures in their distribution. This can cause inspected samples to exhibit artificially elevated directional scores relative to the independently calibrated reference threshold, resulting in false positives. As shown in Table~\ref{tab:ablation_preview}, removing the gate leaves macro-TPR virtually unchanged but increases macro-FPR from at most $1.96\%$ to $47.07$--$63.03\%$. This highlights the importance of class-level screening in preventing benign distributional variations from being misinterpreted as poisoning-related shifts.

\paragraph{Sample scoring.}
By default, SCR uses the directional scoring function defined in Eq.~(\ref{eq:sample_score}) to assign a score to each sample for filtering. To evaluate the effectiveness of this design, we replace directional scoring with residual magnitude, $s_i^{\mathrm{mag}}=\|r_i\|_2$, while keeping all other components unchanged. As shown in Table~\ref{tab:ablation_preview}, this replacement leads to a substantial decrease in macro-TPR across all datasets. This result indicates that residual magnitude alone is less effective at identifying poisoned samples. In contrast, measuring alignment with the estimated shared shift improves macro-TPR, demonstrating the effectiveness of the proposed directional sample scoring.

\subsection{Sensitivity Analysis}
\label{sec:sensitivity}
We examine how the clean-reference budget and victim model architecture
affect SCR. Appendix B.2 reports sensitivity to the
class-screening threshold, sample-level threshold, and feature extraction
layer.

\begin{figure}[t]
\centering
\includegraphics[width=\linewidth]{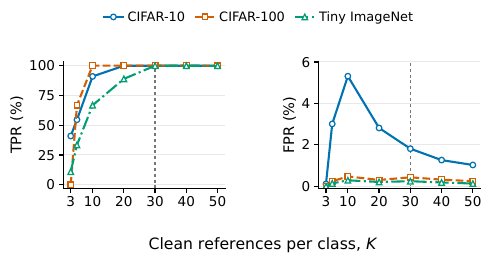}
\caption{Clean-reference budget sensitivity analysis.
Dashed lines mark the default of 30 references per class.}
\label{fig:clean-reference-budget}
\end{figure}

\paragraph{Number of clean samples per class.}
We vary the reference budget over \(K\in\{3,5,10,20,30,40,50\}\), which affects both reconstruction and reference calibration. Very small reference budgets degrade detection due to insufficient coverage of within-class feature variations and unreliable calibration of clean directional scores. Increasing \(K\) progressively improves detection, with TPR largely saturating around \(K=30\), while larger budgets further reduce FPR. These results suggest that 30 references per class provide a favorable balance between detection performance and reference requirements (Fig.~\ref{fig:clean-reference-budget}).

\paragraph{Model architecture.}
We evaluate SCR with ResNet-18, PreAct ResNet-18~\cite{he2016identity}, VGG-16~\cite{simonyan2015very}, and MobileNetV3-Large~\cite{howard2019searching} on the attacks in Table~\ref{tab:main_detection}. We extract flattened features from \texttt{layer4} for both ResNet variants, the last ReLU before the final max-pool in VGG-16, and the final convolutional block before global average pooling in MobileNetV3-Large. Across all architectures and datasets, SCR maintains macro-TPR above $98.9\%$ and macro-FPR below $2.0\%$ (Table~\ref{tab:architecture_sensitivity}), demonstrating that its detection performance is largely insensitive to backbone architecture.

\begin{table}[t]
\centering\scriptsize
\caption{Model architecture sensitivity.}
\label{tab:architecture_sensitivity}
\setlength{\tabcolsep}{2pt}
\renewcommand{\arraystretch}{1.12}
\begin{tabular*}{\columnwidth}{@{\extracolsep{\fill}}lccc@{}}
\toprule
Architecture & CIFAR-10 & CIFAR-100 & Tiny ImageNet \\
\midrule
\textbf{ResNet-18 (default)} & 99.71/1.96 & 99.87/0.43 & 99.81/0.23 \\
PreAct ResNet-18 & 99.53/1.68 & 99.32/0.38 & 99.57/0.21 \\
VGG-16 & 99.05/1.51 & 99.39/0.45 & 99.43/0.29 \\
MobileNetV3-Large & 98.94/1.02 & 99.02/0.51 & 99.48/0.33 \\
\bottomrule
\end{tabular*}
\end{table}

\subsection{Dataset Purification and Retraining}
\label{sec:purification}

We assess clean utility by removing the samples flagged by SCR and
retraining ResNet-18 from scratch on the remaining data. We further add three control methods for comparing the effectiveness of SCR in removing poisoned sample for supressing the attack performance and maintaining clean ACC for downstream training. The semi-oracle
\emph{Random} control matches SCR's removal count in each training class
and selects samples uniformly within that class. \emph{Oracle} removes
all annotated poisons, and \emph{Full Clean} trains on the original
unpoisoned dataset. All runs use the training configuration in
Section~\ref{sec:comparison}. Table~\ref{tab:purification_summary}
reports overall clean ACC, target-class clean ACC ($\mathrm{ACC}_0$),
and ASR.

\begin{table}[t]
\centering\scriptsize
\caption{Purification and retraining (\%) experimental results.}
\label{tab:purification_summary}
\setlength{\tabcolsep}{2pt}
\renewcommand{\arraystretch}{1.08}
\begin{tabular*}{\columnwidth}{@{\extracolsep{\fill}}lrrr@{}}
\toprule
Strategy & ACC $\uparrow$ & ASR $\downarrow$ & $\mathrm{ACC}_0\uparrow$ \\
\midrule
\multicolumn{4}{l}{\textbf{CIFAR-10}} \\
Random & 94.58 & 93.56 & 92.55 \\
Oracle & 94.87 & 1.11 & 95.26 \\
\textbf{SCR} & 94.61 & 0.54 & 92.95 \\
Full Clean & 94.97 & -- & 92.80 \\
\midrule
\multicolumn{4}{l}{\textbf{CIFAR-100}} \\
Random & 74.34 & 66.52 & 74.78 \\
Oracle & 74.62 & 0.20 & 89.89 \\
\textbf{SCR} & 74.75 & 0.08 & 86.33 \\
Full Clean & 74.54 & -- & 87.00 \\
\midrule
\multicolumn{4}{l}{\textbf{Tiny ImageNet}} \\
Random & 62.87 & 59.59 & 67.56 \\
Oracle & 62.98 & 1.12 & 88.22 \\
\textbf{SCR} & 63.11 & 0.12 & 85.78 \\
Full Clean & 62.96 & -- & 82.00 \\
\bottomrule
\end{tabular*}
\end{table}

SCR's mean overall clean ACC stays within $0.4$ percentage points of
both Oracle and Full Clean on every dataset. Its mean target-class ACC
exceeds Full Clean on CIFAR-10 and Tiny ImageNet and is within one
percentage point on CIFAR-100. Oracle achieves higher mean target-class
ACC on all three datasets. With class-wise removal counts matched to
Random, SCR achieves higher overall and target-class ACC on each
dataset and keeps mean ASR below $1\%$, compared with roughly $60\%$
or more for Random. SCR provides effective backdoor suppression after
retraining while maintaining overall and target-class clean accuracy
comparable to training using the full clean dataset.

%% file: sec/5_conclusion.tex
\section{Conclusion}
In this paper, we propose SCR that detect training samples poisoned by single-target backdoor attacks. It reconstructs sample representations from clean references and detects coherent residual shifts through anchored mixture modeling, class-level screening, and directional scoring. SCR requires no prior knowledge of the trigger, target class, or poisoning rate. Experiments covering twelve dirty-label and clean-label attacks across CIFAR-10, CIFAR-100, and Tiny ImageNet yield attack-averaged TPR above $99.5\%$ and FPR below $2\%$ on each dataset, with higher macro-TPR than eight representative baselines. Ablations highlight the complementary roles of sample-specific reconstruction and class-level screening. Retraining after purification yields low ASR and mean overall clean ACC within $0.3$ percentage points of Oracle on each dataset, although class-specific accuracy losses remain. Building on these results, future work will explore extending SCR's class-level screening to more general multi-target settings and developing more reference-efficient and adaptive strategies for selecting clean reference samples, further broadening its applicability across diverse datasets and poisoning scenarios.